%% file: main.tex
\documentclass[12pt, final]{l4dc23/l4dc2023}

\title[Constrained Neurosymbolic Models]{Guaranteed Conformance of Neurosymbolic Models to Natural Constraints}

\usepackage{times}
\input{macros}

\author{%
 \Name{Kaustubh Sridhar$^1$ } \Email{ksridhar@seas.upenn.edu}\\
 \Name{Souradeep Dutta$^1$ } \Email{duttaso@seas.upenn.edu}\\
 \Name{James Weimer$^2$ } \Email{james.weimer@vanderbilt.edu}\\
 \Name{Insup Lee$^1$ } \Email{lee@seas.upenn.edu}\\
 \addr $^1$PRECISE Center, University of Pennsylvania, $^2$Vanderbilt University
}

\begin{document}

\maketitle

\begin{abstract}%
Deep neural networks have emerged as the workhorse for a large section of robotics and control applications, especially as models for dynamical systems. Such data-driven models are in turn used for designing and verifying autonomous systems. 
They are particularly useful in modeling medical systems where data can be leveraged to individualize treatment. 
In safety-critical applications, it is important that the data-driven model is conformant to established knowledge from the natural sciences. Such knowledge is often available or can often be distilled into a (possibly black-box) model. 
For instance, an F1 racing car should conform to Newton's laws (which are encoded within a unicycle model). 
In this light, we consider the following problem - given a model $M$ and a state transition dataset, we wish to best approximate the system model while being a bounded distance away from $M$. We propose a method to guarantee this conformance. Our first step is to distill the dataset into a few representative samples called memories, using the idea of a growing neural gas. Next, using these memories we partition the state space into disjoint subsets and compute bounds that should be respected by the neural network in each subset. This serves as a symbolic wrapper for guaranteed conformance. We argue theoretically that this only leads to a bounded increase in approximation error; which can be controlled by increasing the number of memories. We experimentally show that on three case studies (Car Model, Drones, and Artificial Pancreas), our constrained neurosymbolic models conform to specified models (each encoding various constraints) with order-of-magnitude improvements compared to the augmented Lagrangian and vanilla training methods.\footnote{Our code can be found at: \url{https://github.com/kaustubhsridhar/Constrained_Models}}
\end{abstract}


\input{sections/introduction}
\input{sections/related_work}

\input{sections/problem_formulation}

\input{sections/methods}

\input{sections/preliminaries}

\input{sections/approximating_model_constraints}
\input{sections/error_analysis}

\input{sections/training_with_constraints}
\input{sections/results}

\bibliography{references}
\newpage
\input{sections/appendix}

\end{document}

%% file: macros.tex
\usepackage{amsfonts}
\usepackage{xcolor}
\usepackage{amssymb}
\usepackage{mathtools}
\usepackage{dsfont}

\usepackage{graphicx}

\usepackage{natbib}
\newcommand{\reals} {\mathbb{R}}
\usepackage{amsmath,amsfonts,bm,bbm}

\newtheorem{problem}{Problem Statement}[section]

\usepackage{graphicx}
\usepackage{soul}
\usepackage{algorithm}
\usepackage{algorithmic} 
\usepackage{multirow}
\usepackage{array}
\usepackage{xcolor, colortbl}
\usepackage{wrapfig}

\newcommand{\D}{\mathcal{D}}

\usepackage{tikz}

\input{math_commands}

%% file: math_commands.tex
\usepackage{amsmath,amsfonts,bm}

\def\eqref#1{equation~\ref{#1}}

\def\1{\bm{1}}

\DeclareMathAlphabet{\mathsfit}{\encodingdefault}{\sfdefault}{m}{sl}
\SetMathAlphabet{\mathsfit}{bold}{\encodingdefault}{\sfdefault}{bx}{n}

\DeclareMathOperator*{\E}{\mathbb{E}}

\newcommand{\sigmoid}{\sigma}



%% file: sections/introduction.tex
\section{Introduction}

Deep neural networks (DNNs) are capable of learning highly-complex relationships between input data and the expected output. This permits training and validation of large models in robotics and medicine  \citep{djeumou_l4dc2022, kushner2020conformance, ode6_drone_landing}, enabling designers to comfortably achieve small approximation errors. But the caveat that comes with this flexibility is the lack of generalization when pushed outside of the training distribution. We refer to the experiments in \citet{conformal_predictions} as an example. One of the instances it covers corresponds to that of Newton's first law. The neural network dynamics model of a car should predict that, given zero throttle and when at rest, the car should continue to remain at rest. The neural network model trained on real vehicle trajectory data in \citet{autorally} failed to conform to this simple property. A very similar situation happens in the case of the glucose-insulin dynamics model for an artificial pancreas, a device for patients with type-1 diabetes. This property has been studied in \citet{kushner2020conformance}, where it was found that deep neural network models could easily generate predictions that can be fatal for the patient.

However, these challenges are much less prevalent in models which are typically informed by the different scientific disciplines. Examples of this include models based on mechanical properties of robotic systems \citep{vehicle}, aerodynamic properties of drag and lift \citep{all_about_quads}, physiological models of the human body \citep{DallaManModel, sanjian_jim_model_based} and alike. The \emph{advantage} of using models (rather than atomic constraints) is that they encompass a wider range of desirable properties quite naturally. In robotics, it is common to find such high-fidelity physics-engine-based simulators \citep{carla, pybullet, todorov2012mujoco}. In medical applications, examples include artificial pancreas simulators \citep{DallaManModel, sanjian_jim_model_based}. Unfortunately in practice, such models can be of black-box nature, allowing only samples to be observed. Our goal is to use such models to inviscate a deep neural network into conformal behavior.

\input{sections/intro_figure}

In this work, we propose a method that guarantees the satisfaction of natural constraints by constructing a wrapper for the DNN based on symbolic information. This is achieved through a novel neural gas based partitioning technique and estimation of a model $M$'s output ranges. 
Such a guarantee does not come for free, but shows up as a slightly higher approximation error (which can be attributed to the black-box nature of model $M$). Our contributions can be listed as: 1) A novel memory-based method to constrain neural network dynamics models with guarantees. 2) A theoretical guarantee that our memory-based constraining method guarantees conformance with only a bounded increase in approximation error. 3) Results on three case studies demonstrating that we outperform augmented Lagrangian methods for constraint satisfaction by a few orders of magnitude.


\vspace{-4mm}

%% file: sections/intro_figure.tex
\begin{figure}[t!]
    \centering
    \begin{minipage}{0.47\linewidth}
        \centering
        \includegraphics[width=\linewidth]{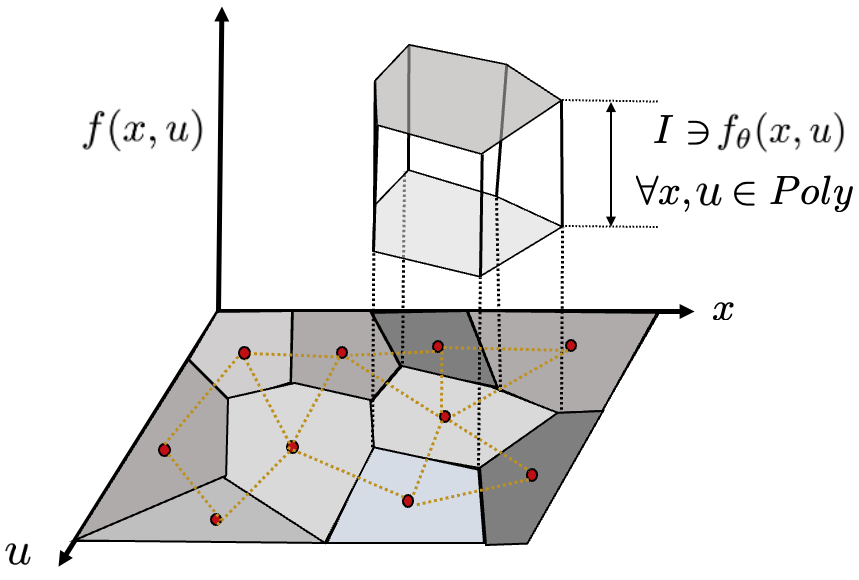}
        \vspace{-5mm}
        \caption{\small Depiction of our neurosymbolic algorithm. First, the input $\mathcal{X}-\mathcal{U}$ plane is partitioned into polyhedrons using a Neural-Gas. For inputs from each polyhedron, we generate sound under-approximations of the model $M$'s output. Next, we learn the dynamics $f_{\theta}$ that is constrained (by construction) to respect these interval constraints. 
        }
        \label{fig:idea}
    \end{minipage}
    \hfill 
    \begin{minipage}{0.48\linewidth}
        \centering
        \includegraphics[width=0.9\linewidth]{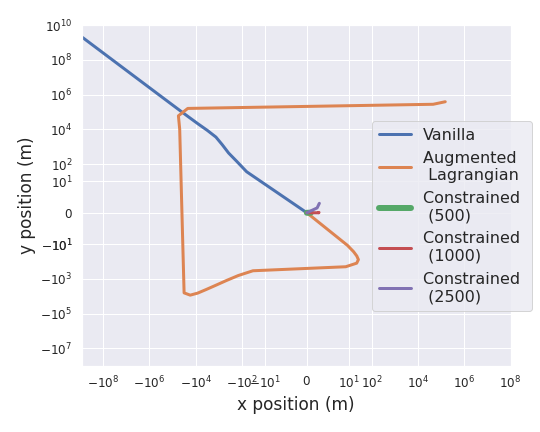}
        \vspace{-5mm}
        \caption{\small Trajectories generated from learned car dynamics models, starting at rest at the origin, with zero control inputs for 20 timesteps. Our neurosymbolic \textit{constrained} models (with varying memories) respect Newton's first law of motion (and remain at rest) unlike \textit{vanilla} and \textit{augmented lagrangian} neural networks that drift away from the origin.}
        \label{fig:test_at_rest}
    \end{minipage}
    \vspace{-8mm}
\end{figure}

%% file: sections/related_work.tex
\section{Related Work}

\noindent\textbf{Enforcing constraints on neural networks}: 
Imposing constraints on deep neural networks has been studied from various perspectives \citep{djeumou_l4dc2022, constraints_neurips, constraints2, constraints3, constraints4, constraints5, constraints6, constraints7, constraints8}. These include constraints of symmetry and contact forces for dynamical systems in \citet{djeumou_l4dc2022}, suitable constraints for specific Lagrangian or Hamiltonian neural networks in \citet{constraints_neurips}, human pose constraints in \citet{constraints2}, path norm constraints on resnets in \citet{constraints3}, partial differential equation (PDE) constraints for inverse design in \citet{constraints4}, Focker-Planck constraints for fusion in \citet{constraints5}, fairness constraints in \citet{constraints6}, constraints on predictive control \citep{sridhar2022predictncritic, zhang2023mpc_recovery}, language label constraints in \citet{constraints7}, and segmentation constraints in \citet{constraints8}. All of these methods rely on the augmented Lagrangian method to train constrained neural networks. Solving the dual problem, \textit{i.e.} converging to a stationary point for the min-max optimization is challenging with neural networks and non-convex constraints \citep{constraints2}. Further, the process is data-hungry and generalizes poorly in out-of-distribution data \citep{conformal_predictions, constraints2, constraints3}. Our focus in this work is to leverage the benefits of the augmented Lagrangian approach (its flexible loss function) but constrain the neural network by design, and with a guarantee, to remain within desirable output bounds computed using models that encode all desired constraints. In the process, we obtain several orders of magnitude reduction in constraint loss and learn with few gradient steps. 

\noindent
\textbf{Physics informed neural networks for dynamics models}: Although our focus is on enforcing constraints, we also briefly discuss related ideas in physics-informed neural networks \citep{pinn, constraints2, lu2021deepxde, physics_prior, lagrangian_nn, hamiltonian_nn}. Physics-informed architectures for dynamical systems in particular have been explored via specific Neural ODE structures for a class of systems \citep{ode1_hamiltonian, ode2, ode3_lagrangian_mechanics, ode4, ode5, ode6_drone_landing} or via a broader Neural ODE structure for a class of vector fields \citep{djeumou_l4dc2022}, all towards learning continuous-time dynamics for robotics applications. Our constraining framework can be applied around any such Neural ODE. But moreover, our constraints can include black-box models and scale quickly to any state and action space unlike NeuralODEs which are restricted to systems with rigorous mathematical models \citep{pinn}. Further, to present a general solution, we make no assumption on the architecture and to extend to applications beyond dynamics models in robotics (such as medicine, computing systems, and operations research), we learn discrete-time dynamics models in our experiments rather than continuous-time dynamics models. 


%% file: sections/problem_formulation.tex
\section{Problem Formulation}

Consider a discrete time non-linear dynamical system $x_{t} = f(x_{t-1}, u_t)$, where $x \in \mathcal{X}$ is the state of the system and $ u \in \mathcal{U}$ is the control input. As a shorthand, we denote $s_t = (x_{t-1}, u_t)$ and we have $s \in \mathcal{S} := \mathcal{X} \times \mathcal{U}$. We can rewrite the unknown discrete-time non-linear map that captures the system dynamics as $f: \mathcal{S} \mapsto \mathcal{X}$ with $x_t = f(s_t)$. We assume access to a dataset $D = {(s_0, x_0), (s_1, x_1),\dots, (s_N, x_{N})}$ drawn from distribution $\mathcal{D}$, such that $x_{t} = f(s_t)$. 
Usually, the goal is to estimate $f$ with a function $f_\theta $, where $\theta \in \reals^p$ is potentially the parameters of a neural network. Typically, the goal of an algorithm which estimates $\theta$ is usually to reduce approximation error on the training dataset $D$. In addition to this, sometimes it is desirable that the estimated model $f_\theta$ satisfies physics-informed constraints \citep{lagrangian_nn}. Next, we define a few relevant concepts.


\begin{definition}[Model Constraint]
\label{defn:constraint}
Assume a  model  $M: \mathcal{S} \mapsto \mathcal{X} $, and a parameter $\delta$. Then the model constraint $\psi^\delta_{M,f_\theta}: \mathcal{S} \mapsto \reals$ is \emph{True} iff $\psi^\delta_{M,f_\theta}(s) > 0$ where $\psi^\delta_{M,f_\theta}(s) := \delta - || M(s) - f_\theta(s) ||_\infty $. 
\end{definition}
Here we assume $M$ to be Lipschitz continuous with constant $L$. We state our problem  next.

\begin{problem}[Constrained Neural Network]
Find a function $f_{\theta}(.): \mathcal{S} \mapsto \mathcal{X}$ , which minimizes the approximation error on dataset $D$, while satisfying the constraints given by $\psi^\delta_{M,f_\theta}$. That is find $ \theta^* = \underset{\theta}{\mathrm{arg min}} \;\; \frac{1}{N} \sum_{i=1}^{N} \lVert f_\theta(s_i) - x_i \rVert_2$, $\text{ subject to,} \quad \psi^\delta_{M,f_\theta}(s) > 0 $.
\end{problem}



%% file: sections/methods.tex
\section{Overall Approach}


To restate, we want our estimated model $f_\theta$ to approximate our training data while respecting the constraint imposed by the model $M$. We use the following intuition in our approach: if restricted to a small enough input region $\hat{\mathcal{S}}$ the output of the model $M$ can be under-approximated by a set $\mathcal{X}_o$. If we can ensure that the predictions of $f_\theta$ stay within this interval then we can bound the  difference between $f_\theta$ and $M$, as being proportional to the size of the input-region $\hat{\mathcal{S}}$, which improves with finer partitioning of the input space. 
Thus, to summarize our approach, we first partition an input space into small enough input regions and for each sub-region, we estimate an interval under-approximation for the values of $M$ which can satisfy $\psi^\delta_{M,f_\theta}$. Next, we train our function approximator $f_\theta$ to respect these interval constraints in each such sub-region. This is accomplished using a \emph{constraining operator} $\Gamma$ on $f_\theta$. In Section \ref{sec:constraint-computation} we explain a method for computing these sound under-approximations of $M$. Then, in Section \ref{sec:approximation-error}, we explain the constraining operator and bound the approximation error incurred due to this operator. Figure \ref{fig:idea} displays our approach.

%% file: sections/preliminaries.tex
\section{Preliminaries}
\label{sec:preliminaries}

We define the idea of a \emph{neural gas} \citep{growing_neural_gas, incrementally_growing_neural_gas, neural_gas}. From a given set of points embedded in a metric space, a growing neural gas algorithm has the ability to learn important topological relations in the form of a graph of prototypical points. It uses a simple Hebb-like learning rule to construct this graph. 
\begin{definition}[Neural Gas]
Neural Gas $\mathcal{G} := (\mathcal{A}, \mathcal{E})$, is composed of the following two components, 
\begin{enumerate}
    \item A set $\mathcal{A} \subset \mathcal{S}$ of the nodes of a network. Each node $m_i \in \mathcal{A}$ is called a memory in this paper.
    \item A set $\mathcal{E} \subset \{ (m_i, m_j) \in \mathcal{M}^2, i \neq j \}$ of edges among pairs of nodes, which inform about the topological structure of the data. The edges are unweighted.
\end{enumerate}
\end{definition}

The edges in $\mathcal{E}$ preserve the neighborhood relations among the data, and is useful in achieving a Voronoi-like partitioning of the data manifold. The graphical structure of a neural gas makes it much more appealing to algorithmically resolve neighborhood relations. For a given node $m_i$, let us denote $\mathcal{E}^i$ as the set of neighbors of $m_i$ according to $\mathcal{G}$.
For most practical purposes in a control setting, the spaces $\mathcal{S}$ and $\mathcal{X}$ are embedded in Euclidean spaces $\reals^t$, and $\reals^d$ respectively, where $t \geq d$. Where, $t-d$ is the dimension of control input. Let $k$ be the cardinality of $\mathcal{A}$ : $\{m_1, m_2, m_3, \dots , m_k\}$. Then, we can define the Voronoi polyhedron \citep{voronoi}, around a given point $m_i$ in the following fashion.
\begin{definition}[Voronoi Polyhedron]
For a point $m_i$ , the Voronoi polyhedron $\mathcal{S}^i_v \in \mathcal{S}$ can be defined using the Euclidean distance function $d : \mathcal{S} \times \mathcal{S} \mapsto \reals$ as,
$$ \mathcal{S}^i_v = \{ s \in \mathcal{S} |\; d(s, m_i) < d(s, m_j) \;\; \forall j \in \mathcal{E}^i \} $$
\end{definition}
In practice, constructing the Voronoi polyhedron $\mathcal{S}^i_v$ can be achieved in the following way. Given points which are neighbors $m_i$ and $m_j$, it is possible to compute a line segment $l_{ij}$ which connects them. Let us denote the perpendicular bisector of $l_{ij}$ as the linear inequality $H_{ij}(s) > 0$. For any point $s$ which is in the same side of $H_{ij}$ as $m_i$ the inequality holds. The reverse is true for the half space constraint $H_{ji}$. This gives us an algorithm to compute $\mathcal{S}^i_v=\underset{j \in \mathcal{E}^i}{\bigcap} \,\,H_{ij}$. Thus, given a set of $k$ nodes the Voronoi tessellation induces a splitting of the space $\mathcal{S}$ into a set of disjoint sets $\mathcal{S}^1, \mathcal{S}^2, \dots, \mathcal{S}^k$. We drop the subscript $v$ for the rest of the paper. Our guarantees of constraint satisfaction is over the union of these subsets.

%% file: sections/approximating_model_constraints.tex
\section{Approximating Model Constraints}
\label{sec:constraint-computation}
Assume a (relatively small) subset $\mathcal{S}_a \subset \mathcal{S}$, and $M^j(s)$ denote the $j$-th output of the model at input $s$. We wish to compute the interval $ I^j_a :=  [\underset{s \in S_a}{min} M^j(s), \underset{s \in \mathcal{S}_a}{max} M^j(s)]$. Assume that $\forall s \in \mathcal{S}_a$, $f^j_\theta(s) \in I^\prime$, and $ I^\prime \subseteq I^j_a$. Where $I^\prime$ is the interval bound on values of $f^j_\theta$ in $\mathcal{S}_a$. Then $\underset{s \in S_a}{max} |  M^j(s) - f^j_\theta(s) | \leq |I^j_a|$. Now, in practice it is hard to precisely compute the interval $I^j_a$ for black-box models $M$. Meaning that we would resort to estimating the min and max of $M^j$ using sampling based techniques. There exists a stochastic optimization algorithm to estimate the true maxima of a Lipschitz function on a bounded domain \citep{Mladineo1991StochasticMO}. Here we follow a simple sampling based rendition to estimate  $I^j_a$. We denote $[k]$ as the list of numbers from $0 \dots k-1$. Next, we note the following lemma.

\begin{lemma}
\label{lem : function_diff}
Let $g : \reals^t \rightarrow \reals$ be an $L_g$-Lipschitz continuous function on a closed and compact set $S_a$, and $l$ and $u$ be its estimated lower and upper bounds. Then, $\forall z \in [l,u] $,  $ \underset{ s \in \mathcal{S}_a}{max} \,\, | g(s) - z| < L_g |S_a|$ . 
\end{lemma}
\noindent{\textbf{Proof: }}The proof can be found in Appendix \ref{sec:proof_appendix}. 
\\

\noindent With $\mathcal{S}_a \subset \reals^t$, let $l$ and $u$ be the estimated minima and maxima of $M^j$. Thus, if $\forall s \in \mathcal{S}_a$, $f^j_\theta(s) \in [l,u]$, then $\underset{s \in S_a}{max} |  M^j(s) - f^j_\theta(s) | \leq L_{M^j} |S_a|$  . Now, across all dimensions $j \in [d]$, let $L_M = max \; L_{M^j}$ then, $ ||f_\theta(s) - M(s)||_\infty <  L_M |S_a|$. Assume $a^*$, to be the largest partition induced by the neural gas $\mathcal{G}$, then setting $\delta = L_M |S_{a^*}|$ ensures satisfaction of model constraint $\psi^\delta_{M,f_\theta}$ in Definition \ref{defn:constraint}. This bound can be made much tighter in practice if the model $M$ is known in an analytical form. Allowing tight computations of its limits possible using techniques like interval arithmetic and Taylor models \citep{rino}

So, given a set $\mathcal{S}$ and using neural gas $\mathcal{G}$, we have a partitioning of $\mathcal{S} = \underset{i \in [k]}{\bigcup} \mathcal{S}^i$. Let us denote this set of partitions of $\mathcal{S}$ as $\mathcal{P}_{\mathcal{S}} := \{ \mathcal{S}^1, \mathcal{S}^2, \dots, \mathcal{S}^k \}$. Also, for each subset $\mathcal{S}^i$ we can compute range estimate $I_i \subset \mathcal{X}$, which respects the constraint $\psi^\delta_{M, f_\theta}$. In the following discussions, let us refer to this constraint map as $\mathsf{C}_{M, \delta} : \mathcal{P}_\mathcal{S} \mapsto \mathcal{I}^d$. Where,  $\mathcal{I}^d$ is a $d-$dimensional interval in $\reals^d$. For a subset in $\mathcal{P}_{\mathcal{S}}$, $\mathsf{C}_{M, \delta}$ returns the appropriate output range.

%% file: sections/error_analysis.tex
\section{Function Approximation Error}
\label{sec:approximation-error}

In this section we define a constraining operator on a function, and analyze the error encountered in the process. The goal of a constraining operator is to threshold the values of the function to be within certain desirable limits. Assume an interval $I \subset \reals^d$, and value $x \in \mathcal{X}$, then we define a projection in the following fashion along each dimension $i$, $Proj^i_I(x) :=$ $I_l^i \;\text{when}\;  x^i \leq I_l^i $; $I_u^i \;\text{when}\;  x^i \geq I_u^i $; and $x^i$ otherwise.

\begin{definition}[Constraining Operator] \label{def:constraining_operator}
A constraining operator $\Gamma_{\mathcal{P}_\mathcal{S}} : \mathcal{X}^{\mathcal{S}} \rightarrow \mathcal{X}^{\mathcal{S}} $ parameterized by the partition set - $\mathcal{P}_\mathcal{S}$, modifies functions to respect the corresponding interval constraints. For a function $F: \mathcal{S} \mapsto \mathcal{X}$, it can be defined in the following fashion,
$$
 \Gamma_{\mathcal{P}_\mathcal{S}}(F(s)):= Proj_{\mathsf{C}_{M, \delta} (\mathcal{S}_q)}(F(s)) \; \text{where, } s \in \mathcal{S}_q\; \text{and,}\; \mathcal{S}_q \in \mathcal{P}_\mathcal{S}
$$

\end{definition}



Hence, the constraining operator $\Gamma_{\mathcal{P}_\mathcal{S}}$ ensures that our estimated model $f_\theta$ which attempts to approximate the true function $f$, also respects the constraint $\psi^\delta_{M, f_\theta}$. Even though we assume that $f \models \psi^\delta_{f_\theta, M}$, our approximation error in building the map $\mathsf{C}_{M, \delta}$ can affect the model approximation error $|f - f_\theta|$. This however as we show only leads to a bounded cost in approximation error. Which can be reduced by adopting finer partitions in $\mathcal{P}_\mathcal{S}$, that is increasing the nodes $\mathcal{A}$ in the neural gas $\mathcal{G}$.



\begin{theorem}[Approximation Error] \label{theorem}
Assume real and continuous functions $f, f_\theta : \mathcal{S} \rightarrow \mathcal{X}$, $\forall s \in \mathcal{S}$, if $||f_\theta(s) - f(s)||_{\infty} < \epsilon$,  then $||\Gamma_{\mathcal{P}_\mathcal{S}}(f_\theta)(s) - f(s)||_{\infty} < 2 \epsilon + \alpha \underset{\mathcal{S}^k \in \mathcal{P}_\mathcal{S}}{max} \; |\mathcal{S}^k|$, where $\alpha$ is some constant.
\end{theorem}

\noindent{\textbf{Proof: }}Assume a generic input $s \in \mathcal{S}$, and $s \in \mathcal{S}^q $ for some $q \in [|\mathcal{P}_\mathcal{S}|] $. Additionally, let $I^q$ be the interval constraint imposed by $\Gamma_{\mathcal{P}_\mathcal{S}}$ on $f_\theta$ using the map $\mathsf{C}_{M, \delta}$. Since the sets $\mathcal{S}$ and $\mathcal{X}$ are embedded in the real spaces $\reals^t$ and $\reals^d$ respectively, we can analyze the error incurred along each dimension. Also, we drop the subscript and denote the constraining operator as simply $\Gamma$ since the partition remains fixed for the remainder of the results.\\
$x_j$ refers to the $j^{th}$ element of $x$.
\noindent Let us pick a dimension $w \in [t]$, we define the lower correction set $\gamma|_{w,l} : \{ s \;|\; \Gamma(f_\theta)(s)_w \geq f_\theta(s)_w \;\text{and}\; s \in \mathcal{S}^q \}$ . Intuitively, this is the set of points in $\mathcal{S}^q$, which need a correction due to underflow. Let us denote the difference function as $\Delta_{w,l}$,
\begin{equation}
\label{eq:delta_defn}
\Delta_{w,l}(s) :=
\begin{cases}
   \Gamma(f_\theta)(s)_w - f_\theta(s)_w &\text{when}\;  s \in \gamma_{w,l} \cap \mathcal{S}^q\\ 
   0 &\text{when}\;  s \in \mathcal{S}_q \setminus \gamma_{w,l} \\ 

\end{cases}.     
\end{equation}

We can similarly define the upper correction set $\gamma_{w,u} \subseteq \mathcal{S}^q$ and the difference function as $\Delta_{w,u}(s) = f_\theta(x)_w - \Gamma(f_\theta)(s)_w$ for $s \in \gamma_{w,u}  \cap \mathcal{S}^q$ and $0$ for anywhere in $\mathcal{S}^q \setminus \gamma_{w,u}$. 

\noindent Now the following is true, for $s \in \gamma_{w,l}$ : $0 \leq \Delta_{w,l}(s) \leq I^q_{w,l} - \underset{x \in \mathcal{S}^q}{min}\; f_\theta(s)_w$.
This is simply due to the bound respected by $\Gamma(f_\theta)(s)_w$.  Due to very similar reasons the following is true as well : $0 \leq \Delta_{w,u}(s) \leq  \underset{s \in \mathcal{S}^q}{max}\; f_\theta(s)_w - I^q_{w,u}$. 
Next, we wish to bound the following quantity: $|\Gamma(f_\theta)(s)_w - f(s)_w|$. The difference between the constrained function and ground truth. Then, 


\begin{equation*}
    \begin{aligned}
        \Gamma(f_\theta)(s)_w - f(s)_w &= f_\theta(s)_w + \Delta_{w,l}(s)  - \Delta_{w,u}(s) - f(s)_w\\
        &= (f_\theta(s)_w - f(s)_w) + (\Delta_{w,l}(s) - \Delta_{w,u}(s)) 
    \end{aligned}
\end{equation*}
The first equality is simply because $\mathcal{S}_q$ can be expressed as a union of the following disjoint sets  $\{ \gamma|_{w,l} \cap \mathcal{S}_q, \gamma|_{w,u} \cap \mathcal{S}^q, \mathcal{S}^q\setminus(\gamma|_{w,u} \cup \gamma|_{w,l} ) \}$ .Therefore, we can write the following, 
\begin{equation*}
    \begin{aligned}
        \Gamma(f_\theta)(s)_w - f(s)_w &\leq \epsilon + (I^q_{w,l} - \underset{s \in \mathcal{S}^q}{min}\; f_\theta(s)_w)
    \text{}\\
        \Gamma(f_\theta)(s)_w - f(s)_w &\geq - \epsilon -  (\underset{s \in \mathcal{S}^q}{max}\; f_\theta(s)_w - I^q_{w,u} ) 
    \end{aligned}
\end{equation*}

Note, the R.H.S of the above equation is negative. Then using the bound on the upper limit of absolute values, we get the following,\\

\begin{equation}
\label{eq:diff_extreme}
    \begin{aligned}
         |\Gamma(f_\theta)(s)_w - f(s)_w | &\leq  \epsilon + \underbrace{(I^q_{w,l} - \underset{s \in \mathcal{S}^q}{min} f_\theta(s)_w)}_{\geq 0} + \epsilon + \underbrace{(\underset{s \in \mathcal{S}^q}{max} f_\theta(s)_w - I^q_{w,u})}_{\geq 0} \\ 
        &= 2 \epsilon + \bigg(\underset{s \in \mathcal{S}^q}{max} f_\theta(s)_w- \underset{s \in \mathcal{S}^q}{min} f_\theta(s)_w\bigg) - \underbrace{(I^q_{w,u} - I^q_{w,l})}_{\text{ constraining width  }}
    \end{aligned}
\end{equation}

Thus, we can bound $|| \Gamma(f_\theta)(s) - f(s) ||_{\infty}$ in the following fashion, for $s \in \mathcal{S}^q$ :
\vspace{-2mm}
\begin{equation*}
    \begin{aligned}
        || \Gamma(f_\theta)(s) - f(s) ||_{\infty} &\leq 2 \epsilon + \underset{w \in [d]}{max} \Bigg(\bigg(\underset{s \in \mathcal{S}^q}{max} f_\theta(s)_w- \underset{s \in \mathcal{S}^q}{min} f_\theta(s)_w\bigg) - \underbrace{(I^q_{w,u} - I^q_{w,l})}_{=|I^q|_w\geq 0} \Bigg) \\ 
        &\leq 2 \epsilon + \underset{w \in [d]}{max} \big( L_{\theta,w} |\mathcal{S}^q| \big) = 2 \epsilon + |\mathcal{S}^q| \underset{w \in [d]}{max} \big( L_{\theta,w}  \big)
    \end{aligned}
\end{equation*}

Now, setting  $\alpha = L_\theta$, where $L_\theta$ is the global Lipschitz constant of $f_\theta$, we can write, 
\begin{equation*}
    \begin{aligned}
        || \Gamma(f_\theta)(s) - f(s) ||_{\infty} 
        &\leq 2 \epsilon + \alpha \underset{\mathcal{S}^q \in \mathcal{P}_\mathcal{S}}{max}\;|\mathcal{S}^q|, \; \forall s \in \mathcal{S} 
    \end{aligned} \hfill \square
\end{equation*} 


We draw the reader's attention to the following terms in inequality \ref{eq:diff_extreme}: 
$(I^q_{w,l} - \underset{s \in \mathcal{S}^q}{min} f_\theta(s)_w)$ and $(\underset{s \in \mathcal{S}^q}{max} f_\theta(s)_w - I^q_{w,u})$. Similar to Lemma \ref{lem : function_diff} it can be shown that these two terms decrease with the size of the set $\mathcal{S}^q$. In other words, having finer Voronoi partitions decreases the approximation error.

%% file: sections/training_with_constraints.tex
\section{Training a Constrained Neural Network Dynamics Model}
\input{sections/simple_algorithm}

We detail our constraining operator used in practice and our overall algorithm below. The inputs to the algorithm are a state transitions dataset $D$ containing ((state, control), (next state)) pairs $(s, x)$, model $M$, and architecture $f_{\theta}: \mathcal{S} \to \mathcal{X}$. 

\textbf{Algorithm \ref{alg:main_simple}.} First (line 1), we generate the unlabelled $\Omega$ dataset which consists of only inputs to the model $s' \in \mathcal{S}$ by sampling throughout the input space but with particular emphasis on relevant regions in $\mathcal{S}$. Next (lines 2, 3), we use the unsupervised neural gas algorithm \citep{neural_gas, growing_neural_gas} to obtain the memories. We partition the input space into voronoi cells around each memory (line 4). With model $M$, we obtain the upper and lower limits along each dimension of the output space (line 5). Finally, we can train the constrained neural network given below,
\begin{align}
    \Gamma(f_{\theta})(s) = \texttt{Lo}(s) + \sigmoid\left(f_{\theta}(s)\right) \, (\texttt{Up}(s) - \texttt{Lo}(s)) \label{eq:constrained_model}
\end{align}
where  $\sigma(.)$ is sigmoid.
Our loss function is the augmented Lagrangian \citep{constraints4} with label and constraint supervison on $\D$ but only constraint supervison on $\Omega$ (where $\psi^\delta_{M, \Gamma(f_\theta)}$ is from Def. \ref{defn:constraint}, $\lambda_i, \mu_i \in \reals$). Additional details of the algorithm can be found in Appendix \ref{sec:algorithm_appendix}.
\vspace{-1mm}
{\small 
\begin{align}
    Loss(\theta, \lambda_1, \mu_2, \lambda_2, \mu_2) = &\E_{\substack{s \sim \mathcal{D} \\ s'\sim \Omega}} \; \bigg[ L\big(\Gamma(f_\theta)(s), x \big) + \bigg( \lambda_1 \psi^\delta_{M, \Gamma(f_\theta)}(s) + \lambda_2 \psi^\delta_{M, \Gamma(f_\theta)}(s') \nonumber \\
    &+ \mu_1 \, \mathds{1}_{(\lambda_1 > 0 \vee \psi > 0)}\, (\psi^\delta_{M, \Gamma(f_\theta)}(s))^2 + \mu_{2} \, \mathds{1}_{(\lambda_2 > 0 \vee \psi > 0)}\, (\psi^\delta_{M, \Gamma(f_\theta)}(s'))^2 \bigg)\bigg] \label{eq:aug_lagrangian}
\end{align} }
\vspace{-10mm}

%% file: sections/simple_algorithm.tex
\begin{algorithm}[t!]
\caption{\texttt{Training a Constrained Neurosymbolic Dynamics Model}}
\small
\flushleft
\textbf{Input:} Dataset $\mathcal{D} = \{(s, x)_i\}_{i \in [N_{\mathcal{D}}]}$, model $M$, DNN architecture $f_{\theta}(.)$ \\ 
\textbf{Output:} Constrained neurosymbolic dynamics model $\Gamma(f_{\theta})(.)$\\
\textbf{Parameters:} Number of memories $n_{\text{memories}}$
\begin{algorithmic}[1]
\STATE Generate input samples $\Omega = \{(s')_i\}_{i \in [N_{\Omega}]}$ \hfill // Generating unlabelled dataset for conformance.
\STATE $D\vert_{inputs}=\{(s)_i\}_{i \in [N_{\mathcal{D}}]} \cup \{(s')_i\}_{i \in [N_{\Omega}]}$ \hfill // combine inputs in both datasets
\STATE Memories $\mathcal{A}$, edges $\mathcal{E}$ $\leftarrow$ NeuralGas($D\vert_{inputs}$, $n_{\text{memories}}$) \hfill // topology of input space
\STATE $\mathcal{S}^1, ..., \mathcal{S}^j, ...$ $\leftarrow$ VoronoiCells($\mathcal{A}, \mathcal{E}$) \hfill // partitions in input space
\STATE Compute lower and upper bounds $I^j_{low}$, $I^j_{up}$ for each partition $\mathcal{S}^j$ using $M$. These are the limits $\texttt{Lo}(s), \texttt{Up}(s)$ for each sample $s$ belonging to a partition $\mathcal{S}^j$.
\STATE Train model $\Gamma(f_{\theta})(.)$ in Equation \ref{eq:constrained_model} with wrapper limits above and self-supervised loss in Equation \ref{eq:aug_lagrangian}.
\end{algorithmic}
\label{alg:main_simple}
\end{algorithm}

%% file: sections/results.tex
\begin{figure}[t]
    \centering
    \includegraphics[width=0.85\linewidth]{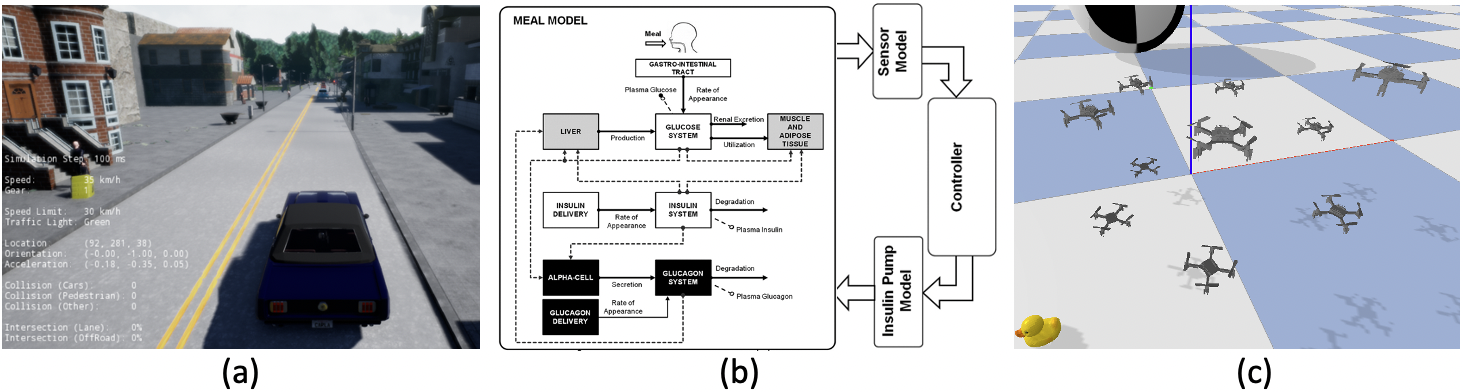}
    \caption{
    Depictions of high-fidelity simulators used in experiments: (a) CARLA \citep{carla}, (b) UVA/Padova Artifical Pancreas \citep{DallaManModel}, (c) Pybullet Drones \citep{gym-pybullet-drones}.
    \vspace{-2em}
    }
    \label{fig:case_studies}
\end{figure}

\begin{figure}[t]
    \centering
    \includegraphics[width=0.9\linewidth]{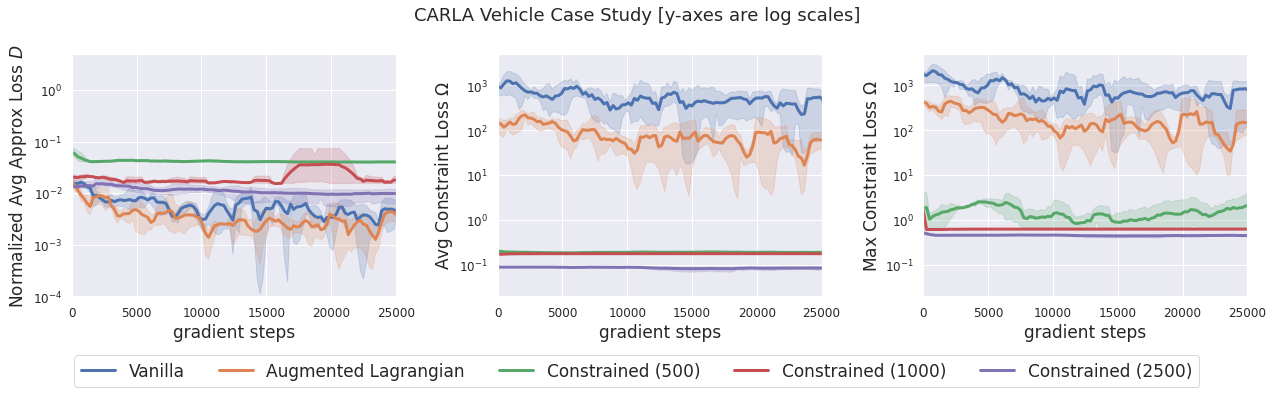}
    \caption{
    \vspace{-1em}
    Plots of approximation loss on $\D$, average constraint loss on $\Omega$, and  maximum constraint loss on $\Omega$ (for 3 random seeds) against steps for
    the CARLA Vehicle case study.
    \vspace{-1.25em}
    }
    \label{fig:results}
\end{figure}

\section{Experiments}
\textbf{Overview and baseline: }
We perform simulated experiments on three case studies. We create a dataset $D$ from high-fidelity simulators that can closely represent reality in each case study. These are depicted in Figure \ref{fig:case_studies}.
Our baseline is the augmented Lagrangian method which utilizes the loss function in \eqref{eq:aug_lagrangian} but uses a standard parameterization $f_{\theta}(.)$ rather than the constrained model given in \eqref{eq:constrained_model}. The augmented Lagrangian method lacks guarantees on constraint satisfaction with deep neural networks and non-convex constraints. We observe that augmented Lagrangian in fact fails to achieve conformance on in-distribution transitions in the test set. 


\textbf{Case Study 1: CARLA -- Conformance of a vehicle model to unicycle dynamics with emphasis on at-rest condition.}
In the first case study, we collect trajectories of x position, y position, heading, velocity, yaw rate from the CARLA simulator \citep{carla, codit} on a variety of terrains and environments (See Figure \ref{fig:case_studies}(a)) for our $\mathcal{D}$ dataset. With previous work \cite{conformal_predictions} having demonstrated the difficulty of learning a dynamics model that predicts no change in state when a vehicle is at rest, we uniformly sample at-rest data for the augmenting dataset $\Omega$. Unicycle dynamics \citep{vehicle, checkpointing_unicycle} are chosen as the model $M$. This implicitly encodes the at-rest condition. We have 15,000 training points, 2000 test points in each of $\D$ and $\Omega$. We select 500, 1000 and 2500 memories to observe the performance with increasing partitions in the training distribution.


We observe, in Figure \ref{fig:results}, that the approximation loss for constrained methods is either similar to or slightly higher than the Vanilla and augmented Lagrangian. This is expected in light of  Theorem \ref{theorem}.
The average constrained loss and max constrained loss on the augmenting dataset $\Omega$ are significantly improved, \textit{by 4 and 3 orders of magnitude respectively} for our method in comparison to Vanilla and augmented Lagrangian.
Moreover, with increasing memories, the constraint loss, both average and maximum on $\Omega$, improve consistently. We also notice that constrained training is highly data-efficient, learning in less than 300 gradient steps unlike the 12000 required by the Augmented Lagrangian.
In Figure \ref{fig:test_at_rest}, we analyze each of the models' predictions starting from the origin at rest, and given zero control inputs for 20 timesteps. We clearly observe that both Vanilla and augmented Lagrangian models predict large drift to the top-left
Constrained models, on the other hand, accurately predict little to no movement. This is also observed at a different random seed in  \ref{fig:test_at_rest_both}.


\begin{figure}[t]
    \centering
    \includegraphics[width=0.485\linewidth]{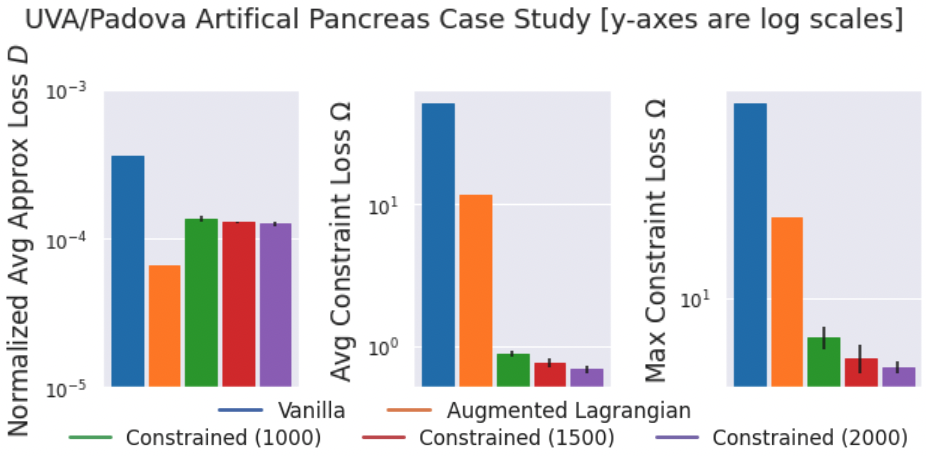}
    \hfill 
    \includegraphics[width=0.485\linewidth]{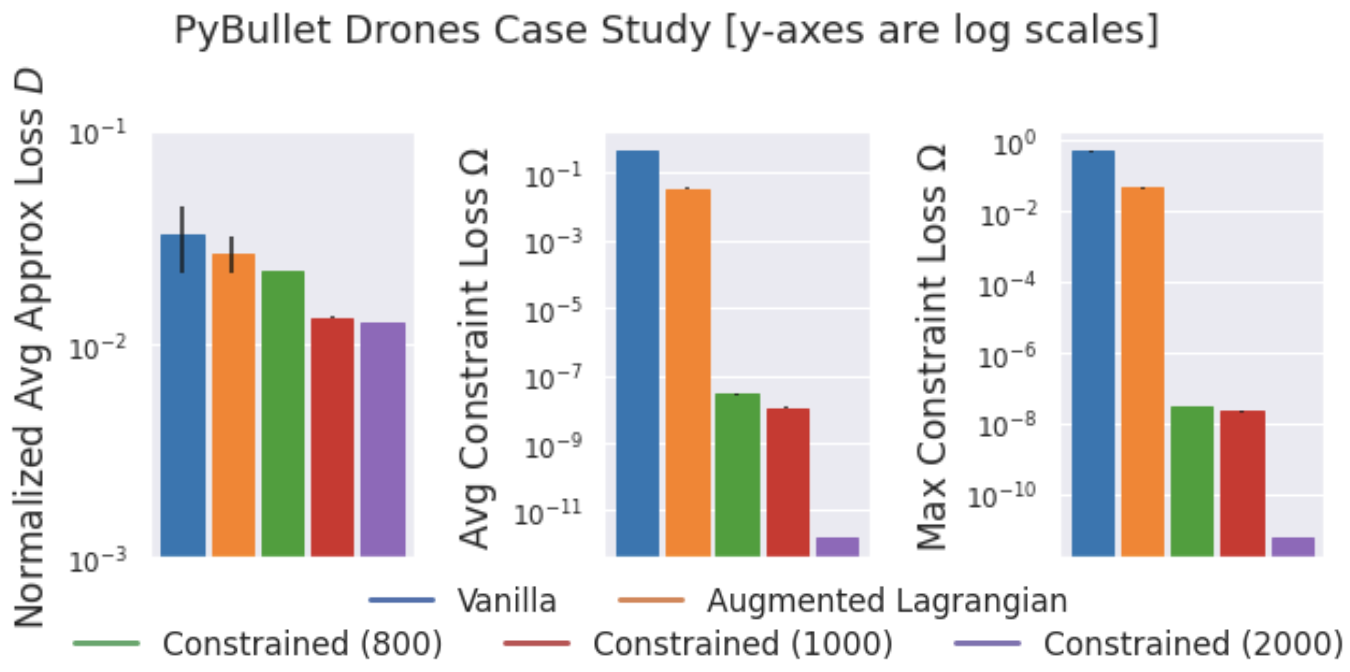}
    \caption{
    \vspace{-0.5em}
    Bar charts of approximation loss on $\D$, average constraint loss on $\Omega$, and maximum constraint loss on $\Omega$ (for 3 random seeds) after training completes for
    AP \textit{(left)}, and Drones \textit{(right)}. 
    Plots of these metrics vs gradient steps for both case studies are in Appendix \ref{sec:plots_appendix}.
    \vspace{-2em}
    }
    \label{fig:bars}
\end{figure}

\textbf{Case Study 2: Artificial Pancreas (AP) -- Conformance of AP models to ARMAX model that encodes glucose-insulin constraints.} We collect traces of glucose, insulin and meal quantities for a patient with the UVA/Padova simulator (See Figure \ref{fig:case_studies}(c)) \citep{DallaManModel} to create the $\D$ dataset. The states consist of a 30 elements-- 10 historical values of glucose, insulin and meals respectively. The model is expected to predict the glucose 5 steps in the future. Each timestep spans $5$ minutes. The intial value of glucose and carbohydrates are randomly chosen in $[150, 190], \,[50,150]$ respectively.

\begin{wraptable}{r}{7cm}
    \scriptsize
    \centering
    \begin{tabular}{|c|c|c|}
        Method & Max. & Avg. \\ 
          &  violation &  violation \\
        \hline
        Vanilla & 3.8356 & 1.315\\
        Aug. Lagrangian & 3.8072 & 1.245\\
        Constrained (1k) & 0.9157 & 0.0092\\
        Constrained (1.5k) & 0.2047 & 0.0027\\
        Constrained (2k) & 0.1775 & 0.0026\\
    \end{tabular}
    \caption{Delta-monotonicity analysis of ``increasing insulin, decreases glucose'' violation in AP models on subsets of test data.}
    \vspace{-5mm}
    \label{tab:AP_analysis}
\end{wraptable}
We also uniformly sample the state space with emphasis on low glucose initial values in $[120, 150]$ and low carbohydrates to create the $\Omega$ dataset. We have 18,750 training points, 2500 test points in each of $\D$ and $\Omega$ datasets. Moreover, for our model $M$, we train a constrained ARMAX model such that any increase in insulin, will reduce glucose. This is accomplished by constraining insulin weights to be negative in the ARMAX model. In Figure \ref{fig:bars}, we observe that approximation loss on $\D$ is similar across all methods with a slight advantage in the favour of our constrained training. Yet, constrained neural networks outperform vanilla and Lagrangian by an order of magnitude in conforming to the ARMAX model on the $\Omega$ and $\D$ datasets.

The delta-monotonicity property of such models in \citep{kushner2020conformance}, refers to the following - everything else remaining fixed, increasing insulin should lead to reduction in blood glucose prediction. In order to test this property we increase the insulin value in each input trace of test set by a random amount in $[0.6, 1.0]$ and observe the prediction. We report this in Table \ref{tab:AP_analysis}. We observe that vanilla and Lagrangian models violate the constraint by a large margin, whereas constrained models increase the prediction by nearly zero amount.

\textbf{Case Study 3: PyBullet Drones -- Conformance of drone models to quadrotor dynamics with emphasis on hover.}
We collect circular flight trajectories of $6$ drones (See Figure \ref{fig:case_studies}(b)) with aerodynamics effects (drag, downwash, ground effect) included in the Pybullet Drones environment \citep{gym-pybullet-drones} to create the $\D$ dataset. The states consist of 20 items -- x, y, z positions and velocities; roll, pitch, yaw and their rates; quaternions, and rpms of each of the four motors. The controls consist of 4 rpm commands. Our model $M$ is given by the quadrotor dynamics \citep{all_about_quads, quads}. For emphasis on hover, we uniformly sample states across the state distribution and uniformly sample controls for balancing gravity (and hence hovering in-place) to create the $\Omega$ dataset. We have 15,000 training points, 2000 test points in each of $\D$ and $\Omega$. We vary the number of memories from 800, 1000, to 2000. Similar to CARLA, we see (in Figure \ref{fig:bars}) that approximation loss on $\D$ is similar across all methods but there is \textit{upto a 6 order-of-magnitude decrease} in the average and maximum constraint loss on $\Omega$ with our constrained training algorithm. We also observe a rather large increase in performance from 1000 to 2000 memories.
We also plot the average constraint loss on $\D$ for all case studies in Appendix \ref{sec:plots_appendix}.
\vspace{-2mm}

\section{Conclusion}
 We demonstrate how DNN training can be constrained using symbolic information which enforces adherence to natural laws.
We report experiments on three case studies where our method achieves many-fold reductions in constraint loss when compared to the augmented Lagrangian. In future work, we plan to create safety-constrained neurosymbolic policies.\\

\noindent\textbf{Acknowledgements} 
This work was supported in part by ARO W911NF-20-1-0080 and AFRL and DARPA FA8750-18-C-0090. Any opinions, findings, conclusions or recommendations expressed in this material are those of the authors and do not necessarily reflect the views of the Air Force Research Laboratory (AFRL), the Army Research Office (ARO), the Defense Advanced Research Projects Agency (DARPA), the Department of Defense, or the United States Government. Additionally, we would like to thank Prof Eric Eaton from the University of Pennsylvania for valuable discussions on a closely related idea.

%% file: sections/appendix.tex
\newpage 
\section*{Appendix}

\subsection{Proof of Lemma 4} \label{sec:proof_appendix}
\begin{lemma}
Let $g : \reals^t \rightarrow \reals$ be an $L_g$-Lipschitz continuous function on a closed and compact set $S_a$, and $l$ and $u$ be its estimated lower and upper bounds. Then, $\forall z \in [l,u] $,  $ \underset{ s \in \mathcal{S}_a}{max} \,\, | g(s) - z| < L_g |S_a|$ . 
\end{lemma}
\noindent{\textbf{Proof : }} Note that $g$ is a real and continuous function on the connected set $\mathcal{S}_a$ in the metric space $\reals^t$. Since, there exists points $s_l$ and $s_u$ which map to $l$ and $u$ respectively, then by Theorem $4.22$ \cite{Rudin}, for any $ z \in [l , u]$
there exists $s_z \in \mathcal{S}_a$  such that $z = g(s_z)$.  Then we can write the following :
$\underset{s \in S_a}{max} \,\, |g(s) - g(s_z) | \leq L_g |s - s_z| \leq L_g |S_a|$. This completes the proof.

\subsection{Detailed Algorithm for Training a Constrained Neural Network Dynamics Model}
\label{sec:algorithm_appendix}
\input{sections/APP_training_with_constraints.tex}

\subsection{Additional Details of Experiments}
For the CARLA vehicle and PyBullet Drones models, we use a two layer MLP with 1024 neurons in each layer. In the UVA/Padova Artificial Pancreas case study, we use a three layer neural network with 20 neurons in each layer. We utilize the Adam optimizer in all case studies and choose a learning rate with grid search in $[0.001, 0.1]$. We also utilize training batch sizes of 64 for both $\D$ and $\Omega$ datasets. Further, for the CARLA and Drones case studies, we set $\gamma$ to 0. For Artificial Pancreas, we used $\gamma=0.99$.



\subsection{Additional Plots} \label{sec:plots_appendix}

\begin{figure}[H]
    \begin{center}
    \includegraphics[width=0.4\linewidth]{Figures/predictions_at_rest_0_seed_20timesteps.png}
    $\;\;\;\;\;\;\;$
    \includegraphics[width=0.4\linewidth]{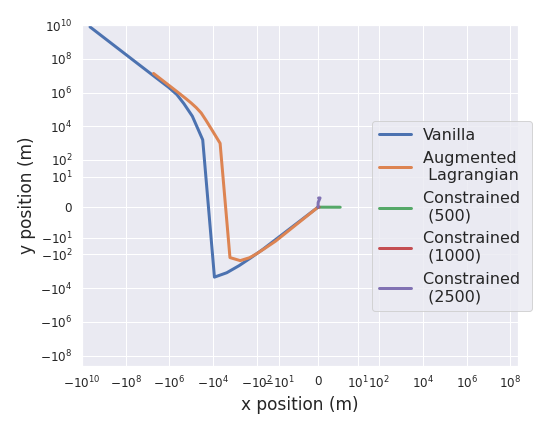}
    \end{center}
    \vspace{-8mm}
    \caption{Analysis of CARLA model prediction drift starting from origin at rest when given zero control inputs for 20 timesteps for random seed of 0 \textit{(left)} and random seed of 1 \textit{(right)}.}
    \label{fig:test_at_rest_both}
\end{figure}

\begin{figure}[ht!]
    \centering
    \includegraphics[width=\linewidth]{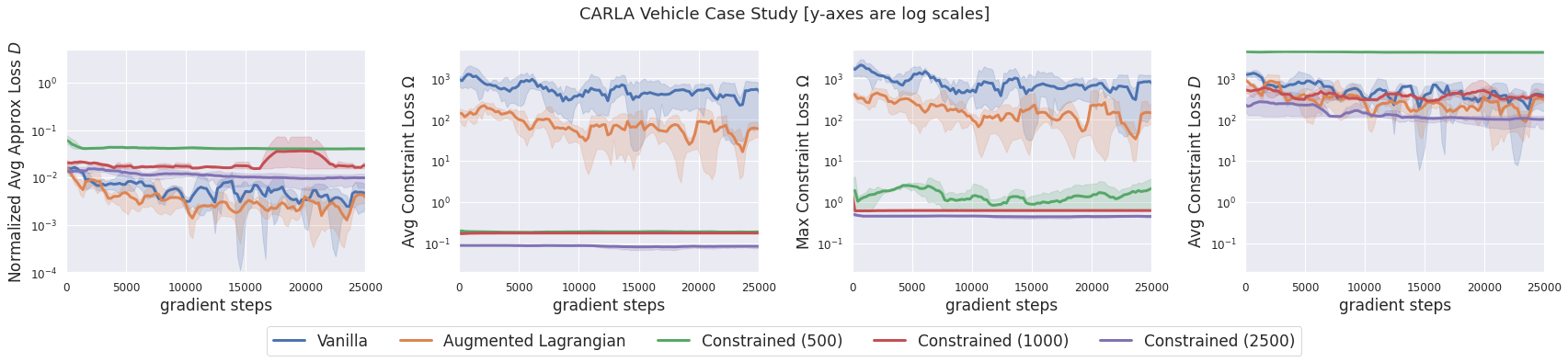}
    \includegraphics[width=\linewidth]{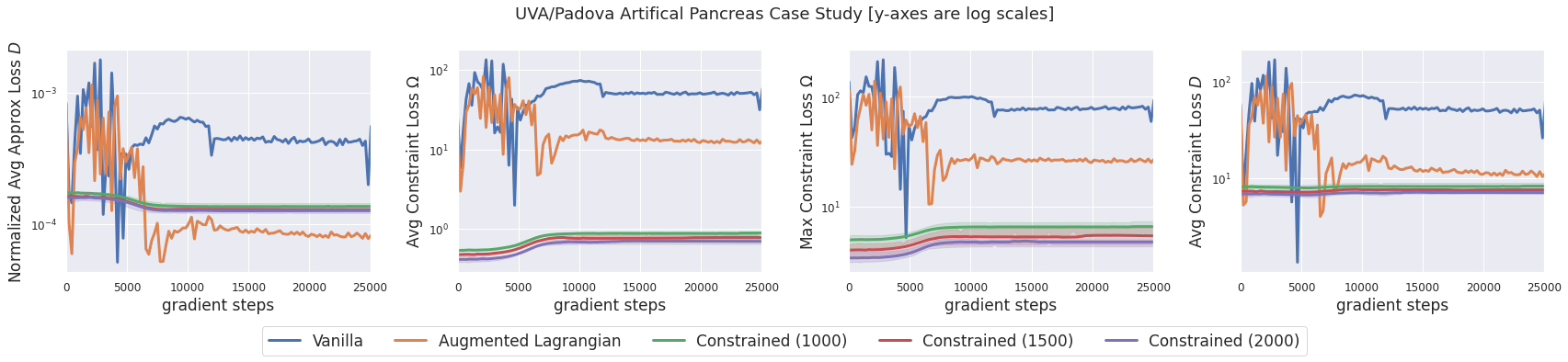}
    \includegraphics[width=\linewidth]{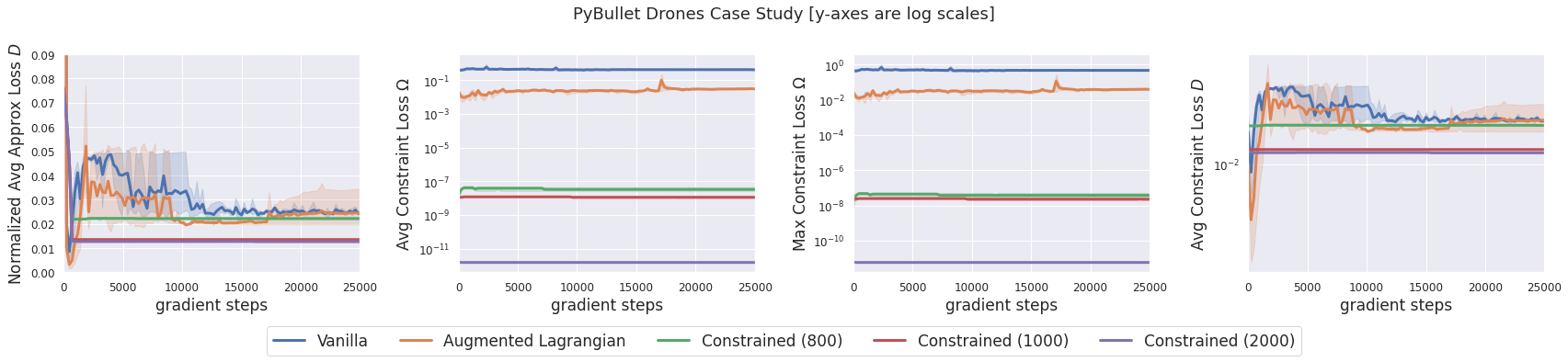}
    \vspace{-5mm}
    \caption{Plots of approximation loss on $\D$, average constraint loss on $\Omega$, maximum constraint loss on $\Omega$, and average constraint loss on $\D$ (for 3 random seeds) against gradient steps for
    the CARLA Vehicle \textit{(top row)}, Artifical Pancreas \textit{(second row)}, and PyBullet Drones \textit{(third row)} case studies.
    }
    \label{fig:results_4}
\end{figure}

%% file: sections/APP_training_with_constraints.tex
\begin{algorithm}[H]
\caption{\texttt{Training a Constrained Neural Network Dynamics Model}}
\small
\flushleft
\textbf{Input:} Dataset $\mathcal{D} = \{(s, x)_i\}_{i \in [N_{\mathcal{D}}]}$, Knowledge $M$, DNN architecture $f_{\theta}(.)$ \\ 
\textbf{Output:} Constrained neural network dynamics model $\Gamma(f_{\theta})(.)$\\
\textbf{Parameters:} Number of memories $n_{\text{memories}}$, batch sizes $N_{\D_{\text{batch}}}, N_{\Omega_{\text{batch}}}$, $0\leq \gamma<1$, N\_Steps, update\_freq
\begin{algorithmic}[1]
\STATE Generate input samples $\Omega = \{(s')_i\}_{i \in [N_{\Omega}]}$ \hfill // unlabelled dataset
\STATE $D\vert_{inputs}=\{(s)_i\}_{i \in [N_{\mathcal{D}}]} \cup \{(s')_i\}_{i \in [N_{\Omega}]}$ \hfill // combine inputs in both datasets
\STATE Memories $\mathcal{A}$, edges $\mathcal{E}$ $\leftarrow$ NeuralGas($D\vert_{inputs}$, $n_{\text{memories}}$) \hfill // topology of input space
\STATE $\mathcal{S}^1, ..., \mathcal{S}^j, ...$ $\leftarrow$ VoronoiCells($\mathcal{A}, \mathcal{E}$) \hfill // partitions in input space
\FOR{each voronoi cell $S^j$} 
    \STATE Sample points inside the cell, propagate through model $M$, and compute lower and upper bounds $I_{low}^j = \min_{s \sim \mathcal{S}^j} M(s)$ and $I_{up}^j = \max_{s \sim \mathcal{S}_j} M(s)$
\ENDFOR 
\FOR{$s$ in $D$, $\Omega$} 
    \STATE $S_j$ $\leftarrow$ FindVoronoiCell($s$, $\mathcal{A}$)\\
    \STATE  Set $\texttt{Lo}(s)$, $\texttt{Up}(s)$ $\leftarrow$ $I_{low}^j, I_{up}^j$ \hfill // output bounds for datasets
\ENDFOR
\FOR{step in N\_Steps}
    \STATE Sample batches $\D_{\text{batch}} = \text{Sample}(D, N_{\D_{\text{batch}}})$, $\Omega_{\text{batch}} = \text{Sample}(D, N_{\Omega_{\text{batch}}})$
    \STATE Set $\texttt{Lo}(s, \text{ step}) = \texttt{Lo}(s) - \gamma^{\text{step}} \; (\texttt{Up}(s) - \texttt{Lo}(s))$ and $
    \texttt{Up}(s, \text{ step}) = \texttt{Lo}(s) + \gamma^{\text{step}} \; (\texttt{Up}(s) - \texttt{Lo}(s))$
    \STATE Compute $\Gamma(f_{\theta})(.)$ for $\D_{\text{batch}}$ and $\Omega_{\text{batch}}$ using $\texttt{Lo}(s, \text{ step})$ and $\texttt{Up}(s, \text{ step})$ \hfill // constrained DNN (\ref{eq:constrained_model_appendix})
    \STATE Compute $Loss(\theta, \lambda, \mu)$ \hfill // augmented Lagrangian loss (\ref{eq:aug_lagrangian_appendix}) or vanilla approximation loss
    \STATE $\theta \leftarrow Optimization\_Step(Loss, \theta, \D_{\text{batch}}, \Omega_{\text{batch}})$ 
    \IF{step \% update\_freq == 0}
    \STATE $\lambda_1, \lambda_2, \mu_1, \mu_2 \leftarrow Update\_Step(\psi^\delta_{M, \Gamma(f_\theta)}, \lambda, \mu)$
    \ENDIF
\ENDFOR 
\RETURN $\Gamma(f_{\theta})(.)$
\end{algorithmic}
\label{alg:main}
\end{algorithm}

We detail our Algorithm in this section. The inputs to the algorithm are a state transitions dataset $D$ containing ((state, control), (next state)) pairs $(s, x)$, model $M$, and architecture $f_{\theta}: \mathcal{S} \to \mathcal{X}$. 

\textbf{Algorithm \ref{alg:main}} First (line $1$), we generate the unlabelled $\Omega$ dataset which consists of only inputs to the model $s' \in \mathcal{S}$ by sampling throughout the input space but with particular emphasis on relevant regions in $\mathcal{S}$. Then, (lines $2-4$), we use the unsupervised neural gas algorithm \citep{neural_gas, growing_neural_gas} to obtain the neural gas graph $\mathcal{G} = (\mathcal{A}, \mathcal{E})$. We utilize these memories and edges, to create partitions of the input space as voronoi cells with memories at their center. In each voronoi cell, we sample points, propagate them through the model $M$ and obtain the upper and lower limits along each dimension of the output space $\mathcal{X}$ (lines 5-7). This creates the constraint map $\mathsf{C}$. Using this, we can find the lower and upper bounds of each point in $\D$ and $\Omega$ (lines 8-12). First, we locate the corresponding voronoi cell, and then use the bounds computed in Line $6$. Finally, we can train the constrained neural network (denoted $\Gamma(f_{\theta})(.)$) as follows,
\begin{align}
    \Gamma(f_{\theta})(s) = \texttt{Lo}(s) + \sigmoid\left(f_{\theta}(s)\right) \, (\texttt{Up}(s) - \texttt{Lo}(s)) \label{eq:constrained_model_appendix}
\end{align}
where $f_{\theta}: \mathcal{S} \to \mathcal{X}$ is a parameterized function which maps from the input space to output space, and $\sigma(x): \mathcal{X} \to [0, 1]$ is the sigmoid function.
Equation \ref{eq:constrained_model_appendix} is but one realization of the constraining operator discussed in Definition \ref{def:constraining_operator}. Our loss function is the augmented Lagrangian loss \citep{constraints4} itself and is given below $\big($where $\psi^\delta_{M, \Gamma(f_\theta)}(s) = \delta - ||M(s) - \Gamma(f_{\theta})(s)||\big)$.
{\small 
\begin{align}
    Loss(\theta, \lambda_1, \mu_2, \lambda_2, \mu_2) = &\E_{\substack{s \sim \mathcal{D} \\ s'\sim \Omega}} \; \bigg[ L\big(\Gamma(f_\theta)(s), x \big) + \bigg( \lambda_1 \psi^\delta_{M, \Gamma(f_\theta)}(s) + \lambda_2 \psi^\delta_{M, \Gamma(f_\theta)}(s') \nonumber \\
    &+ \mu_1 \, \mathds{1}_{(\lambda_1 > 0 \vee \psi > 0)}\, (\psi^\delta_{M, \Gamma(f_\theta)}(s))^2 + \mu_{2} \, \mathds{1}_{(\lambda_2 > 0 \vee \psi > 0)}\, (\psi^\delta_{M, \Gamma(f_\theta)}(s'))^2 \bigg)\bigg] \label{eq:aug_lagrangian_appendix}
\end{align} }

We can then train the neural network by back-propagating through the constrained neural network (lines 12-16). We enhance gradient feedback under constrained outputs with an exponential schedule on the lower and upper bounds (line 13). We also intermittently update the slack variables through a schedule or as a gradient ascent step on the value of the constraint $\psi^\delta_{M, \Gamma(f_\theta)}$ (lines 17-19).